\documentclass{article}

\PassOptionsToPackage{numbers,square,comma,sort}{natbib}
\usepackage[preprint]{neurips_2023}

\usepackage[utf8]{inputenc} 
\usepackage[T1]{fontenc}    
\usepackage{url}            
\usepackage{booktabs}       
\usepackage{amsfonts}       
\usepackage{nicefrac}       
\usepackage{microtype}      
\usepackage{xcolor}         
\usepackage{amssymb}
\usepackage{float}
\usepackage{amsmath}
\usepackage{amsthm}
\usepackage{graphicx}
\usepackage{diagbox}
\usepackage{tabularx}
\usepackage{bm}
\usepackage{bbm}
\usepackage{algorithm}
\usepackage{algorithmic}
\usepackage{setspace}
\usepackage{subfigure}

\usepackage{amsfonts}
\usepackage{float}
\usepackage{enumitem,graphicx}
\usepackage{wrapfig}
\usepackage{multirow}
\usepackage{xcolor}
\usepackage[pagebackref=true,breaklinks=true,colorlinks,bookmarks=false]{hyperref}

\def\w{\bm{w}}
\def\x{\bm{x}}

\def\vtheta{\bm{\theta}}
\def\vdelta{\bm{\delta}}

\def\gL{\mathcal{L}}

\makeatletter
\DeclareRobustCommand\onedot{\futurelet\@let@token\@onedot}
\def\@onedot{\ifx\@let@token.\else.\null\fi\xspace}

\def\eg{\emph{e.g.}} 

\def\ie{\emph{i.e.}}

\def\wrt{\emph{w.r.t.}} 

\def\etal{\emph{et al.}}

\long\def\comment#1{}
\newcommand*{\email}[1]{%
    \normalsize\href{mailto:#1}{#1}\par
    }

\usepackage{times}
\usepackage{epsfig}
\usepackage{graphicx}
\usepackage{amsmath}
\usepackage{amssymb}
\usepackage{dsfont}
\usepackage{xcolor}

\def\vtheta{{\bm{\theta}}}

\def\gL{{\mathcal{L}}}

\def\gW{{\mathcal{W}}}
\def\gX{{\mathcal{X}}}
\def\gY{{\mathcal{Y}}}

\DeclareMathOperator*{\argmax}{arg\,max}
\DeclareMathOperator*{\argmin}{arg\,min}

\title{Neural Polarizer:  A Lightweight and Effective Backdoor Defense via Purifying Poisoned Features}
\author{
Mingli Zhu\textsuperscript{1,2}\thanks{These authors contributed equally to this work.}
\ \ \ \ 
Shaokui Wei\textsuperscript{1,2$\ast$} \ \ \ \ 
Hongyuan Zha\textsuperscript{1} \ \ \ \ 
Baoyuan Wu\textsuperscript{1,2}\thanks{Corresponds to Baoyuan Wu (\email{wubaoyuan@cuhk.edu.cn}).} \\
\textsuperscript{1}School of Data Science, The Chinese University of Hong Kong, Shenzhen \\
(CUHK-Shenzhen), China \\
\textsuperscript{2}Shenzhen Research Institute of Big Data 
}

\begin{document}

\maketitle

\begin{abstract}
Recent studies have demonstrated the susceptibility of deep neural networks to backdoor attacks. Given a backdoored model, its prediction of a poisoned sample with trigger will be dominated by the trigger information, though trigger information and benign information coexist. Inspired by the mechanism of the optical polarizer that a polarizer could pass light waves with particular polarizations while filtering light waves with other polarizations, we propose a novel backdoor defense method by inserting a learnable neural polarizer into the backdoored model as an intermediate layer, in order to purify the poisoned sample via filtering trigger information while maintaining benign information. The neural polarizer is instantiated as one lightweight linear transformation layer, which is learned through solving a well designed bi-level optimization problem, based on a limited clean dataset. Compared to other fine-tuning-based defense methods which often adjust all parameters of the backdoored model, the proposed method only needs to learn one additional layer, such that it is more efficient and requires less clean data. Extensive experiments demonstrate the effectiveness and efficiency of our method in removing backdoors across various neural network architectures and datasets, especially in the case of very limited clean data.

\end{abstract}

\section{Introduction\label{sec1}}

Several studies have revealed the vulnerabilities of deep neural networks (DNNs) to various types of attacks \cite{moosavi2016deepfool,kurakin2018adversarial,ilyas2018black,gao2022imperceptible}, of which backdoor attacks \cite{gu2019badnets,cheng2023tat,wang2023robust} are attracting increasing attention.
In backdoor attacks, the adversary could produce a backdoored DNN model through manipulating the training dataset \cite{chen2017targeted,Trojannn} or the training process \cite{nguyen2021wanet, li2021invisible}, such that the backdoored model predicts any poisoned sample with particular triggers to the predetermined target label, while behaves normally on benign samples. Backdoor attacks can arise from various sources, such as training based on a poisoned dataset, or utilizing third-party platforms for model training, or downloading backdoored models from untrusted third-party providers. These scenarios significantly elevate the threat of backdoor attacks to DNNs' applications, and meanwhile highlight the importance of defending against backdoor attacks.

Several seminal backdoor defense methods have been developed, mainly including 1) in-training approaches, which aim to train a secure model based on a poisoned dataset through well designed training algorithms or objective functions, such as DBD \cite{huang2022backdoor} and D-ST \cite{chen2022effective}.
2) post-training approaches, which aim to mitigate the backdoor effect from a backdoored model through adjusting the model parameters (\eg, fine-tuning or pruning), usually based on a limited subset of clean training dataset, such as fine-pruning \cite{liu2018fine}, ANP \cite{wu2021adversarial}, or i-BAU \cite{zeng2022adversarial}. 
This work focuses on the latter one. 
However, there are two limitations to existing post-training approaches. First, given very limited clean data, it is challenging to find a good checkpoint to simultaneously achieve  backdoor mitigation and benign accuracy maintenance from the high-dimensional loss landscape of a complex model. Second, adjusting all parameters of a complex model is costly. 

\begin{figure}
\centering
\includegraphics[width=0.9\textwidth]{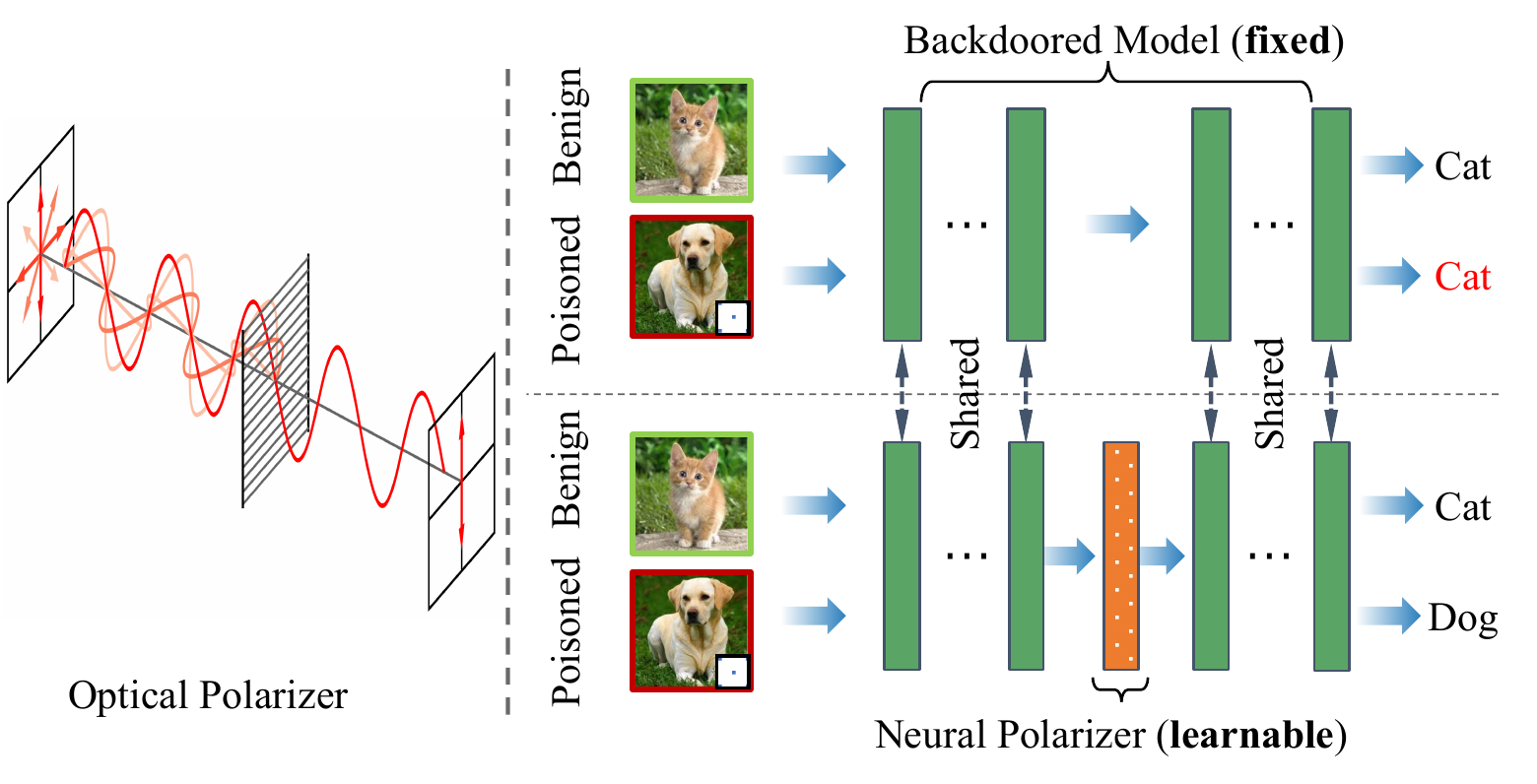}
\caption{\textbf{Left}: Illustration of an optical polarizer \cite{Polarizer}. Only light waves with specific polarizations can pass through the polarizer among the three incident light waves. \textbf{Right}: Defense against backdoors by integrating a trainable neural polarizer into the compromised model. The neural polarizer effectively filters out backdoor-related features, effectively eliminating the backdoor.}
\label{fig: motivation}
\end{figure}

To tackle the above limitations, we propose a lightweight and effective post-training defense approach, which only learns one additional layer, while fixing all layers of the original backdoored model. 
It is inspired by the mechanism of the optical polarizer \cite{xiong2016optical} that in a mixed light wave with diverse polarizations, only the light wave with some particular polarizations could pass the polarizer, while those with other polarizations are blocked (see Fig. \ref{fig: motivation}-left).  
Correspondingly, by treating one poisoned sample as the mixture of trigger feature and benign feature, we define a neural polarizer and insert it into the backdoored model as one additional intermediate layer (see Fig. \ref{fig: motivation}-right) in order to filter trigger feature and maintain benign feature, such that poisoned samples could be purified to mitigate backdoor effect, while benign samples are not significantly influenced.

In practice, to achieve an effective neural polarizer, it should be learned to weaken the correlation between the trigger and the target label while keeping the mapping from benign samples to their ground-truth labels. 
However, the defender only has a limited clean dataset, while neither trigger nor target label is accessible. To tackle it, we propose a bi-level optimization, where the target label is estimated by the output confidence, and the trigger is approximated by the targeted adversarial perturbation.  
Besides, in our experiments, the neural polarizer is implemented by a linear transformation (\ie, the combination of one $1\times1$ convolutional layer one batch normalization layer). Consequently, it can be efficiently and effectively learned with very limited clean data to achieve good defense performance, which is verified by extensive experiments on various model architectures and datasets.

Our main contributions are three-fold. \textbf{1)} We propose an innovative backdoor defense approach that only learns one additional linear transformation called neural polarizer while fixing all parameters of the backdoored model, such that it just requires very low computational cost and very limited clean data. 
\textbf{2)} A bi-level formulation and an effective learning algorithm are provided to optimize the parameter of the neural polarizer. 
\textbf{3)} Extensive experimental results demonstrate the superiority of the proposed method on various networks and datasets, in terms of both efficiency and effectiveness.

\section{Related work}
\paragraph{Backdoor attack and defense.}
Traditional backdoor attacks are additive attacks that modify a small fraction of training samples by patching a pre-defined pattern and assigning them to targeted labels \cite{gu2019badnets, chen2017targeted}. These modified samples, along with the unaffected samples, constitute a \textit{poisoned dataset} \cite{gu2019badnets}. The model trained with this dataset will be implanted with backdoors that predict the targeted label when triggered by the injected patterns, while maintaining normal behavior on clean samples \cite{gu2019badnets, chen2017targeted}. Recently, advanced attacks have considered more invisible trigger injection methods such as training an auto-encoder feature embedding or using a local transformation function \cite{zeng2021rethinking, li2021invisible, nguyen2021wanet}. To increase the stealthiness of attacks, clean-label attacks succeed by obfuscating image subject information and establishing a correlation between triggers and targeted labels without modifying the labels of poisoned samples \cite{shafahi2018poison, barni2019new}.

Backdoor defense methods can be broadly categorized into training-stage defenses and post-processing defenses. Training-stage defenses assume access to a poisoned dataset for model training. The different behaviors between the poisoned and clean samples can be leveraged to identify suspicious instances, such as sensitivity to transformation \cite{chen2022effective} and clustering phenomenon in feature space \cite{huang2022backdoor}. Most defense methods belong to post-processing defenses, which assume that the defender only has access to a suspicious DNN model and a few clean samples. Therefore, they must remove the backdoor threat with limited resources. There are mainly three types of defense strategies: trigger reversion methods try to recover the most possible triggers and utilize the potential triggers to fine-tune the model \cite{wang2019neural}; pruning-based methods aim at locating the neurons that are most related to backdoors and pruning them \cite{liu2018fine, zheng2022data, wu2021adversarial,zheng2022pre}; and fine-tuning based defenses leverage clean samples to rectify the model \cite{li2021neural, zeng2022adversarial}. I-BAU \cite{zeng2022adversarial} is most closely related to our method, which formulates a minimax optimization framework to train the network with samples under universal adversarial perturbations. However, our method differs from I-BAU in that our approach does not require training the entire network, and our perturbation generation mechanism is distinct. Other methods proposed for backdoor detection include Beatrix \cite{ma2022beatrix}, which uses Gram matrices to identify poisoned samples; and AEVA \cite{guo2021aeva}, which detects backdoor models by adversarial extreme value analysis. In this study, we focus on post-processing defenses and primarily compare our method with state-of-the-art post-processing defenses \cite{wang2019neural, liu2018fine, li2021neural}.

\paragraph{Adversarial training.}
Adversarial training (AT) \cite{madry2017towards,bai2021recent,andriushchenko2020understanding} is an effective technique to improve the robustness of DNNs by incorporating adversarial examples during training. One of the most well-known methods is PGD-AT \cite{madry2017towards}, which searches for adversarial examples by taking multiple iterative steps in the direction of maximizing the loss function. AWP \cite{wu2020adversarial} proposes a method to improve the robustness of models by introducing adversarial perturbations in the weights of the network. MART \cite{wang2020improving} proposes a new misclassification-aware adversarial training method by studying the impact of misclassified examples on the robustness of AT. Feature Denoising (FD) is an adversarial defense technique that further improves the robustness of adversarial training by applying denoising-based operations. CIIDefence \cite{gupta2019ciidefence} presents a defense mechanism against adversarial attacks by searching for pixel-level perturbations around the original inputs that can cause high-confidence misclassifications. HGD \cite{liao2018defense} proposes an adversarial defense mechanism that uses a high-level representation-guided denoiser to remove perturbations from the input image. Xie \etal \cite{xie2019feature} introduce a denoising block that uses convolutions and residual connections to denoise feature maps using non-local means. In adversarial training  literature, the most related work to ours is FD. However, we remark that our method differs from FD in two perspectives. Firstly, FD inserts multiple denoising layers with a residual structure to the network, which is different from ours. Given the limited training samples, multiple denoising layers will cause instability in the network. Secondly,  FD primarily aims to enhance adversarial robustness through self-supervised learning, encoder-decoder reconstruction, or similar techniques \cite{shao2022open}, while  we employ an explicit target label strategy to enhance the robustness against backdoors specifically.

\section{Methodology\label{sec3}}

\subsection{Basic settings\label{sec3.1}}
\paragraph{Notations.} We consider a classification problem with $K$ classes ($K\geq 2$). Let $\x\in \gX \subset \mathbb{R}^d$ be a $d$-dimensional input sample, and its ground truth label is denoted as $y\in \gY = \{1,\dots,K\}$.  Then, a $L$ layers deep neural network $f:\gX\times\gW\to \mathbb{R}^K$ parameterized by $\w\in\gW$ is defined as:
\begin{equation}
f(\x ; \w)=f^{(L)}_{\w_{L}} \circ f^{(L-1)}_{\w_{L-1}} \circ \cdots \circ f^{(1)}_{\w_{1}} (\x),
\end{equation}
where $f^{(l)}_{\w_{l}}$ is the function (\eg, convolution layer) with parameter $\w_{l}$ in the $l^{\text {th }}$ layer with $1 \leq l \leq L$. For simplicity, we denote $f(\x ; \w)$ as $f_{\w}(\x)$ or $f_{\w}$. Given input $\x$, the predicted label of $\x$ is given by $\argmax_k f_k(\x;\w), k=1,\ldots,K$,  where $f_k(\x; \w)$ is the logit of the $k^{\text {th}}$ class.

\paragraph{Threat model.} 
We assume that the adversary could produce the backdoored model $f_{\w}$ through manipulating training data or training process, such that $f_{\w}$ performs well on benign sample $\x$ (\ie, $f_{\w}(\x) = y$), and predicts poisoned sample $\x_{\Delta} = r(\x, \Delta)$ to the target label $T$, with $\Delta$ indicating the trigger and $r(\cdot, \cdot)$ being the fusion function of $\x$ and $\Delta$. 
Considering that the adversary may set multiple targets, we use $T_i$ to denote the target label for $\x_i$.

\paragraph{Defender's goal.}
Assume that the defender has access to the backdoored model $f_{\w}$ and a limited set of benign training data, denoted as $\mathcal{D}_{bn}=\{(\x_i,y_i)\}_{i=1}^{N}$. 
The defender's goal is to obtain a new model $\hat{f}$ based on $f_{\w}$ and $\mathcal{D}_{bn}$, such that the backdoor effect will be mitigated and the benign performance is maintained in $\hat{f}$.

\begin{wrapfigure}{r}{0.39\textwidth}
\centering
\includegraphics[width=0.39\textwidth]{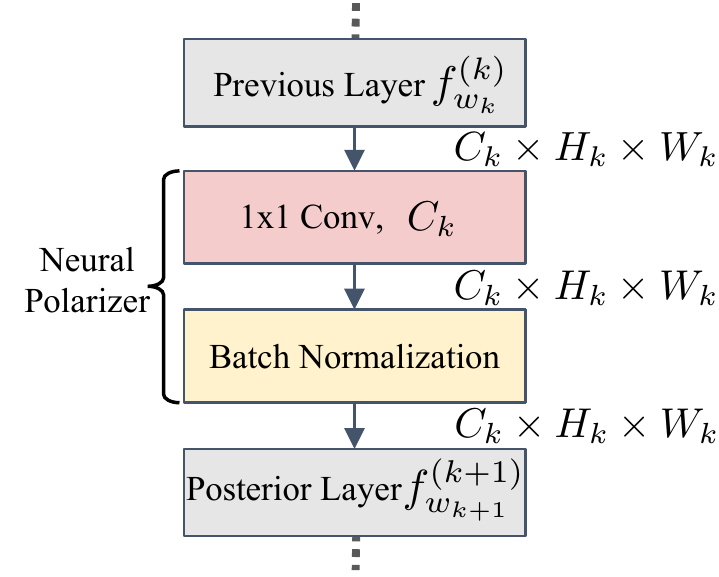}
 \vspace{-1.5em}
\caption{An example of neural polarizer for a DNN.}
\vspace{-0.1in}
\label{fig:insert}
\end{wrapfigure}

\subsection{Neural polarizer \label{sec3.2} for DNNs}

\paragraph{Neural polarizer.} We propose the neural polarizer to purify poisoned sample in the feature space. Formally, it is instantiated as a lightweight linear transformation $g_{\vtheta}$, parameterized with $\vtheta$. As shown in Fig. \ref{fig: motivation}, $g_{\vtheta}$ is inserted into the neural network $f_{\w}$ at a specific immediate layer, to obtain a combined network $f_{\w, \vtheta}$. For clarity, we denote $f_{\w, \vtheta}$ as $\hat{f}_\vtheta$, since $\w$ is fixed. 
A desired neural polarizer should have the following three properties:
\begin{itemize}[leftmargin=*]
    \item \textbf{Compatible with the neighboring layers}: in other words, its input feature and its output activation must have the same shape. This requirement can be fulfilled through careful architectural design.
    \item \textbf{Filtering trigger features in poisoned samples}: after the neural polarizer, the trigger features should be filtered, such that the backdoor is deactivated, \ie, $\hat{f}_\vtheta(\x_{\Delta})\neq T$.
    \item \textbf{Preserving benign features in poisoned and benign samples}: the neural polarizer should preserve benign features, such that $\hat{f}_\vtheta$ performs well on both poisoned and benign samples, \ie, $\hat{f}_\vtheta(\x_{\Delta})= \hat{f}_\vtheta(\x) = y$. 
\end{itemize}
The first property could be easily satisfied by designing neural polarizer's architecture. For example, as illustrated in Fig. \ref{fig:insert}, given the input feature map with the shape $C_k \times H_k \times W_k$, the neural polarizer is implemented by a convolution layer (Conv) with $C_k$ convolution filters of shape $1\times1$, followed by a Batch Normalization (BN) layer. The Conv-BN block can be seen as a linear transformation layer. 
To satisfy the latter two properties, $\vtheta$ should be learned by solving some well designed optimization, of which the details is presented in Section \ref{sec3.3}.

\subsection{Learning neural polarizer\label{sec3.3}}

\paragraph{Loss functions.} 
To learn a good neural polarizer $g_{\vtheta}$, we firstly present some loss functions to encourage $g_{\vtheta}$ to satisfy the latter two properties mentioned above, as follows: 
\begin{itemize}[leftmargin=*]
    \item \textbf{Loss for filtering trigger features in poisoned samples.} Given trigger $\Delta$ and target label $T$, filtering trigger features can be implemented by weakening the connection between $\Delta$ and $T$ with the following loss: 
    \begin{equation}
        \gL_{asr}({\x,y,\Delta,T;\vtheta}) = -\log(1- s_{T}(\x_{\Delta};\vtheta)),
    \end{equation}
    where $s_{T}(\x_{\Delta};\theta)$ is the softmax probability of predicting $\x_{\Delta}$ to label $T$ by $\hat{f}_{\vtheta}$. 
    By reducing $\gL_{asr}$, the attack success rate  of the backdoor attack can be decreased.
    \item \textbf{Loss for maintaining benign features in poisoned samples.} To purify the poisoned sample such that it can be classified to the true label, we leverage the boosted cross entropy defined in \cite{wang2020improving}:
    \begin{equation}
        \gL_{bce}({\x,y,\Delta;\vtheta}) = -\log(s_{y}(\x_{\Delta};\vtheta))-\log\big(1- \max_{k\neq y} s_{k}(\x_{\Delta};\vtheta)\big).
        \label{eq: bce loss}
    \end{equation}
    \item \textbf{Loss for maintaining benign features in benign samples.} 
    To preserve benign features in benign sample, we adopt the widely used cross-entropy loss, \ie, $\gL_{bn}({\x,y;\vtheta}) = \gL_{\text{CE}}(\hat{f}_{\vtheta}(\x),y).$
\end{itemize}

\paragraph{Approximating $\Delta$ and $T$.}
Note that as $\Delta$ and $T$ are inaccessible in both $\gL_{asr}$ and $\gL_{bce}$, these two losses cannot be directly optimized. Thus, we have to approximate $\Delta$ and $T$. 
In terms of approximating $T$, although some methods have been developed to detect the target class of a backdoored model $f_{\w}$ \cite{ma2022beatrix,guo2021aeva}, here we adopt a simple and dynamic strategy that $T \approx y' = \argmax_{k\neq y} \hat{f}_k(\x;\vtheta)$.  
Then, the trigger is approximated by the targeted adversarial perturbation of $f_{\w}$, \ie, 
\begin{equation}\label{eq:at}
\Delta \approx \vdelta^* = \argmin_{\|\vdelta\|_p\leq \rho} \gL_{\text{CE}}\left(\hat{f}_{\vtheta}(\x+\vdelta), y'\right),
\end{equation}
where $\|\cdot\|_p$ is the $L_p$ norm, $\rho$ is the budget for perturbations.
\paragraph{Bi-level optimization.}
Combining the above loss functions and the approximations of $\Delta$ and $T$, we introduce the following bi-level minimization problem to learn $\vtheta$ based on the clean data $\mathcal{D}_{bn}$: 
\begin{equation}\label{loss}
\begin{aligned}
    \min_{\vtheta} \quad \quad & \frac{1}{N}\sum_{i=1}^N \lambda_1 \gL_{bn}({\x_i,y_i};\vtheta) +  \lambda_2 \gL_{asr}({\x_i,y_i,\vdelta^*_i, y'_i;\vtheta}) + \lambda_3 \gL_{bce}({\x_i,y_i,\vdelta^*_i;\vtheta}),\\
    \mathrm{s.t.} \quad \quad & \vdelta^*_i = \argmin_{\|\vdelta_i\|_p\leq \rho} \gL_{\text{CE}}\left(\hat{f}_{\vtheta}(\x_i+\vdelta_i), y'_i\right), 
    ~
    y'_i = \argmax_{k_i\neq y_i} \hat{f}_{k_i}(\x_i;\vtheta), 
\end{aligned}
\end{equation}
where $\lambda_1,\lambda_2,\lambda_3 > 0 $ are hyper-parameters to adjust the importance of each loss function. 

To solve the above optimization problem, we proposed Algorithm \ref{alg:denoise}, dubbed Neural Polarizer based backdoor Defense (\textbf{NPD}). Specifically, NPD solves problem (\ref{loss}) by  alternatively updating the surrogate target label $y'$, the perturbation $\vdelta$ and $\vtheta$ as follows:
\begin{itemize}[leftmargin=*]
    \item \textbf{Inner minimization:} Given parameter $\vtheta$ of the neural polarizer, we first estimate the target label for sample $\x_i$ by $ y'_i = \argmax_{k_i\neq y_i} \hat{f}_{k_i}(\x_i;\vtheta)$. Then, the targeted Project Gradient Descent (PGD) \cite{madry2017towards} is employed to generate the perturbation $\vdelta_i^*$ via Eq. (\ref{eq:at}).
    \item  \textbf{Outer minimization:} Given $y'$ and $\vdelta^*$ for each sample in a batch, the $\vtheta$ can be updated by taking one stochastic gradient descent \cite{bottou2007tradeoffs} (SGD) step \wrt~ the outer minimization objective in Eq. ($\ref{loss}$).
\end{itemize}

\paragraph{Extension.} 
To comprehensively evaluate the performance of NPD, we also provide two variants, with the relaxation that if the target label $T$ is known. One is that approximating the trigger by the targeted adversarial perturbation for each benign sample in Eq. (\ref{loss}), \ie,  $\Delta \approx \vdelta^*_i = \argmin_{\|\vdelta_i\|_p\leq \rho} \gL_{\text{CE}}\left(f(\x_i+\vdelta_i), T\right)$, dubbed \textbf{NPD-TP}.
The other is approximating the trigger by the targeted universal adversarial perturbation (TUAP) \cite{lu2021exploring} for all benign samples, dubbed \textbf{NPD-TU}.

\begin{algorithm}[h]
\caption{Neural Polarizer based Backdoor Defense (NPD)}\label{alg:denoise}
\begin{algorithmic}[1]
\STATE \textbf{Input:} Training set $\mathcal{D}_{bn}$, backdoored model $f_{\w}$, neural polarizer $g_{\vtheta}$,
learning rate $\eta>0$, perturbation bound $\rho>0$, norm $p$, hyper-parameters $\lambda_1$,$\lambda_2$,$\lambda_3>0$, warm-up epochs $\mathcal{T}_0$, training epochs $\mathcal{T}$.
\STATE \textbf{Output:} Model $\hat{f}(\w,\vtheta)$.
\STATE Initialize $\vtheta$ to be an identity function, fix $\w$, and construct the composed network $\hat{f}(\w,\vtheta)$.
\STATE Warm-up: Train $\hat{f}(\w,\vtheta)$ with $\mathcal{L}_{\text{CE}}(D_{bn})$ for $\mathcal{T}_0$ epochs.
\FOR {$t=0,...,$ $\mathcal{T} -1$}
\FOR {mini-batch $\mathcal{B}=\{(\x_i,y_i)\}_{i=1}^{b}\subset \mathcal{D}_{bn}$}
\STATE Compute $\{y'_i\}_{i=1}^{b}$ and generate perturbations $\{(\boldsymbol{\delta_i})\}_{i=1}^{b}$ with $\boldsymbol{\|\delta_{i}\|_p}\leq \rho$ and $\{y'_i\}_{i=1}^{b}$ by targeted PGD attack \cite{madry2017towards} via the inner minimization of Eq. (\ref{loss});
\STATE Update $\vtheta$ via outer minimization of Eq. (\ref{loss}) by SGD.
\ENDFOR
\ENDFOR
\RETURN Model $\hat{f}(\w,\vtheta)$. 
\end{algorithmic}
\end{algorithm}

\section{Experiments}

\subsection{Implementation details\label{sec4.1}}
\paragraph{Attack settings.} We evaluate the proposed method on eight famous backdoor attacks, including BadNets \cite{gu2019badnets} (BadNets-A2O and BadNets-A2A refer to attacking one target label and all labels, respectively), Blended attack (Blended) \cite{chen2017targeted}, Input-aware dynamic backdoor attack (Input-aware)\cite{nguyen2020input}, Low frequency attack (LF) \cite{zeng2021rethinking}, Sample-specific backdoor attack (SSBA) \cite{li2021invisible}, Trojan backdoor attack (Trojan) \cite{Trojannn}, and Warping-based poisoned networks (WaNet) \cite{nguyen2021wanet}. We follow the default attack conﬁguration as in BackdoorBench \cite{wubackdoorbench} for a fair comparison. The poisoning ratio is set to 10\% in comparison with SOTA defenses. These attacks are conducted on three benchmark datasets: CIFAR-10 \cite{krizhevsky2009learning}, Tiny ImageNet \cite{le2015tiny}, and GTSRB \cite{stallkamp2011german}. We test all attacks on PreAct-ResNet18 \cite{he2016identity} and VGG19-BN \cite{simonyan2014very}.

\paragraph{Defense settings.} We compare the proposed methods with six SOTA backdoor defense methods, \ie, Fine-pruning (FP) \cite{liu2018fine}, NAD \cite{li2021neural}, NC \cite{wang2019neural}, ANP \cite{wu2021adversarial}, i-BAU \cite{zeng2022adversarial}, and EP \cite{zheng2022pre}. All these defenses have access to 5\% benign training samples. The training hyperparameters are adjusted based on BackdoorBench \cite{wubackdoorbench}. We evaluate the proposed method under two proposed scenarios and compare our NPD-TU, NPD-TP, and NPD with SOTA defenses. For the ablation study, we focus solely on NPD, which represents a more generalized scenario.
We apply an $l_2$ norm constraint to the adversarial perturbations, with a perturbation bound of 3 for CIFAR-10 and GTSRB datasets, and 6 for Tiny ImageNet. We train the neural polarizer for 50 epochs with batch size 128 and learning rate $0.01$ on each dataset and the transformation block is inserted before the third convolution layer of the fourth layer for PreAct-ResNet18. The loss hyper-parameters $\lambda_1$,$\lambda_2$,$\lambda_3$ are set to $1,0.4,0.4$ for NPD, and $1,0.1,0.1$ for NPD-TU and NPD-TP. More implementation details on SOTA attacks, defenses, and our methods can be found in Section B of \textbf{supplementary materials}.

\paragraph{Evaluation metric.}
In this work, we use clean ACCuracy (ACC), Attack Success Rate (ASR), and Defense Effectiveness Rating (DER)  as evaluation metrics to assess the performance of different defenses. ACC represents the accuracy of clean samples while ASR measures the ratio of successfully misclassified backdoor samples to the target label. Defense Effectiveness Rating (DER $\in [0,1]$ \cite{zhu2023enhancing}) is a comprehensive measure that considers both ACC and ASR:
\begin{equation}
    \mathrm{DER}=[\max (0, \Delta \mathrm{ASR})-\max (0, \Delta \mathrm{ACC})+1] / 2,
\end{equation}
where $\Delta \text{ASR}$ denotes the decrease of ASR after applying defense, and $\Delta \text{ACC}$ denotes the drop in ACC following defense. 
Higher ACC, lower ASR and higher DER indicate better defense performance. 
Note that in comparison with SOTA defenses, the one achieving the best performance is highlighted in \textbf{boldface}, while the second-best result is indicated by \underline{underlining}.

\begin{table}[ht]
\caption{Comparison with the state-of-the-art defenses on CIFAR-10 dataset with 5\% benign data and 10\% poison ratio on PreAct-ResNet18 (\%).}
\centering
\renewcommand\arraystretch{1.1}
\scalebox{0.65}{%
\begin{tabular}{cccccccccccccccc}
\toprule
\multicolumn{1}{c|}{\multirow{2}{*}{ATTACK}} &
\multicolumn{3}{c|}{Backdoored} &
\multicolumn{3}{c|}{FP \cite{liu2018fine}} &
\multicolumn{3}{c|}{NAD \cite{li2021neural}} &
\multicolumn{3}{c|}{NC \cite{wang2019neural}} &
\multicolumn{3}{c}{ANP \cite{wu2021adversarial}} \\
\multicolumn{1}{c|}{} &
ACC &
ASR &
\multicolumn{1}{c|}{DER} &
ACC &
ASR &
\multicolumn{1}{c|}{DER} &
ACC &
ASR &
\multicolumn{1}{c|}{DER} &
ACC &
ASR &
\multicolumn{1}{c|}{DER} &
ACC &
ASR &
DER \\ \midrule
\multicolumn{1}{c|}{BadNets-A2O \cite{gu2019badnets}} &
91.82 &
93.79 &
\multicolumn{1}{c|}{N/A} &
\textbf{91.77} &
\underline{0.84} &
\multicolumn{1}{c|}{\textbf{96.45}} &
88.82 &
1.96 &
\multicolumn{1}{c|}{94.42} &
90.27 &
1.62 &
\multicolumn{1}{c|}{95.31} &
\underline{91.65} &
3.83 &
94.90 \\
\multicolumn{1}{c|}{BadNets-A2A \cite{gu2019badnets}} &
91.89 &
74.42 &
\multicolumn{1}{c|}{N/A} &
92.05 &
1.31 &
\multicolumn{1}{c|}{86.56} &
90.73 &
1.61 &
\multicolumn{1}{c|}{85.82} &
89.79 &
1.11 &
\multicolumn{1}{c|}{85.60} &
\underline{92.33} &
2.56 &
85.93 \\
\multicolumn{1}{c|}{Blended \cite{chen2017targeted}} &
93.44 &
97.71 &
\multicolumn{1}{c|}{N/A} &
92.74 &
10.17 &
\multicolumn{1}{c|}{93.42} &
92.25 &
47.64 &
\multicolumn{1}{c|}{74.44} &
\textbf{93.69} &
99.76 &
\multicolumn{1}{c|}{50.00} &
\underline{93.45} &
47.14 &
75.28 \\
\multicolumn{1}{c|}{Input-Aware \cite{nguyen2020input}} &
94.03 &
98.35 &
\multicolumn{1}{c|}{N/A} &
94.05 &
1.62 &
\multicolumn{1}{c|}{98.36} &
\textbf{94.08} &
0.92 &
\multicolumn{1}{c|}{\underline{98.72}} &
93.84 &
10.48 &
\multicolumn{1}{c|}{93.84} &
\underline{94.06} &
1.57 &
98.39 \\
\multicolumn{1}{c|}{LF \cite{zeng2021rethinking}} &
93.01 &
99.06 &
\multicolumn{1}{c|}{N/A} &
92.05 &
21.32 &
\multicolumn{1}{c|}{88.39} &
91.72 &
75.47 &
\multicolumn{1}{c|}{61.15} &
\textbf{93.01} &
99.06 &
\multicolumn{1}{c|}{50.00} &
\underline{92.53} &
26.38 &
86.10 \\
\multicolumn{1}{c|}{SSBA \cite{li2021invisible}} &
92.88 &
97.07 &
\multicolumn{1}{c|}{N/A} &
\underline{92.21} &
20.27 &
\multicolumn{1}{c|}{88.06} &
92.15 &
70.77 &
\multicolumn{1}{c|}{62.78} &
\textbf{92.88} &
97.07 &
\multicolumn{1}{c|}{50.00} &
92.02 &
16.18 &
90.02 \\
\multicolumn{1}{c|}{Trojan \cite{Trojannn}} &
93.47 &
99.99 &
\multicolumn{1}{c|}{N/A} &
92.24 &
67.73 &
\multicolumn{1}{c|}{65.52} &
92.18 &
5.77 &
\multicolumn{1}{c|}{96.46} &
91.85 &
51.03 &
\multicolumn{1}{c|}{73.67} &
\textbf{92.71} &
84.82 &
57.21 \\
\multicolumn{1}{c|}{WaNet \cite{nguyen2021wanet}} &
92.8 &
98.9 &
\multicolumn{1}{c|}{N/A} &
92.94 &
\textbf{0.66} &
\multicolumn{1}{c|}{\textbf{99.12}} &
\underline{93.07} &
\underline{0.73} &
\multicolumn{1}{c|}{\underline{99.08}} &
92.80 &
98.90 &
\multicolumn{1}{c|}{50.00} &
\textbf{93.24} &
1.54 &
98.68 \\ \midrule
\multicolumn{1}{c|}{AVG} &
92.92 &
94.91 &
\multicolumn{1}{c|}{N/A} &
\underline{92.51} &
15.49 &
\multicolumn{1}{c|}{89.50} &
91.88 &
25.61 &
\multicolumn{1}{c|}{84.13} &
92.27 &
57.38 &
\multicolumn{1}{c|}{68.44} &
\textbf{92.75} &
23.00 &
85.87 \\ \bottomrule \toprule
\multicolumn{1}{c|}{\multirow{2}{*}{ATTACK}} &
\multicolumn{3}{c|}{i-BAU \cite{zeng2022adversarial}} &
\multicolumn{3}{c|}{EP \cite{zheng2022pre}} &
\multicolumn{3}{c|}{NPD-TU (\textbf{Ours})} &
\multicolumn{3}{c|}{NPD-TP (\textbf{Ours})} &
\multicolumn{3}{c}{NPD (\textbf{Ours})} \\
\multicolumn{1}{c|}{} &
ACC &
ASR &
\multicolumn{1}{c|}{DER} &
ACC &
ASR &
\multicolumn{1}{c|}{DER} &
ACC &
ASR &
\multicolumn{1}{c|}{DER} &
ACC &
ASR &
\multicolumn{1}{c|}{DER} &
ACC &
ASR &
DER \\ \midrule
\multicolumn{1}{c|}{BadNets-A2O \cite{gu2019badnets}} &
87.43 &
4.48 &
\multicolumn{1}{c|}{92.46} &
89.80 &
1.26 &
\multicolumn{1}{c|}{95.26} &
90.81 &
1.44 &
\multicolumn{1}{c|}{95.67} &
90.90 &
\textbf{0.62} &
\multicolumn{1}{c|}{\underline{96.12}} &
88.93 &
1.26 &
94.82 \\
\multicolumn{1}{c|}{BadNets-A2A \cite{gu2019badnets}} &
89.39 &
1.29 &
\multicolumn{1}{c|}{85.32} &
88.72 &
3.00 &
\multicolumn{1}{c|}{84.12} &
91.66 &
\underline{0.82} &
\multicolumn{1}{c|}{\underline{86.68}} &
\textbf{92.54} &
\textbf{0.04} &
\multicolumn{1}{c|}{\textbf{87.19}} &
91.41 &
0.89 &
86.52 \\
\multicolumn{1}{c|}{Blended \cite{chen2017targeted}} &
89.43 &
26.82 &
\multicolumn{1}{c|}{83.44} &
91.94 &
48.22 &
\multicolumn{1}{c|}{74.00} &
91.88 &
\textbf{0.03} &
\multicolumn{1}{c|}{\textbf{98.06}} &
91.33 &
0.83 &
\multicolumn{1}{c|}{97.38} &
91.18 &
\underline{0.41} &
\underline{97.52} \\
\multicolumn{1}{c|}{Input-Aware \cite{nguyen2020input}} &
89.91 &
\underline{0.02} &
\multicolumn{1}{c|}{97.10} &
93.68 &
2.88 &
\multicolumn{1}{c|}{97.56} &
92.01 &
0.14 &
\multicolumn{1}{c|}{98.10} &
93.24 &
\textbf{0.00} &
\multicolumn{1}{c|}{\textbf{98.78}} &
89.57 &
0.11 &
96.89 \\
\multicolumn{1}{c|}{LF \cite{zeng2021rethinking}} &
88.92 &
11.99 &
\multicolumn{1}{c|}{91.49} &
91.97 &
84.73 &
\multicolumn{1}{c|}{56.64} &
91.42 &
\textbf{0.01} &
\multicolumn{1}{c|}{\underline{98.73}} &
91.92 &
\underline{0.08} &
\multicolumn{1}{c|}{\textbf{98.94}} &
90.06 &
0.21 &
97.95 \\
\multicolumn{1}{c|}{SSBA \cite{li2021invisible}} &
86.53 &
2.89 &
\multicolumn{1}{c|}{93.92} &
91.67 &
4.33 &
\multicolumn{1}{c|}{95.76} &
91.61 &
\underline{2.46} &
\multicolumn{1}{c|}{\underline{96.67}} &
91.82 &
\textbf{0.83} &
\multicolumn{1}{c|}{\textbf{97.59}} &
90.88 &
2.77 &
96.15 \\
\multicolumn{1}{c|}{Trojan \cite{Trojannn}} &
89.29 &
0.54 &
\multicolumn{1}{c|}{97.64} &
92.32 &
2.49 &
\multicolumn{1}{c|}{98.18} &
\underline{92.59} &
\underline{0.04} &
\multicolumn{1}{c|}{\textbf{99.54}} &
92.19 &
\textbf{0.00} &
\multicolumn{1}{c|}{\underline{99.36}} &
92.37 &
6.51 &
96.19 \\
\multicolumn{1}{c|}{WaNet \cite{nguyen2021wanet}} &
90.70 &
0.88 &
\multicolumn{1}{c|}{97.96} &
90.47 &
96.52 &
\multicolumn{1}{c|}{50.02} &
92.18 &
3.24 &
\multicolumn{1}{c|}{97.52} &
92.57 &
7.47 &
\multicolumn{1}{c|}{95.60} &
91.57 &
0.80 &
98.43 \\ \midrule
\multicolumn{1}{c|}{Avg} &
88.95 &
6.11 &
\multicolumn{1}{c|}{92.42} &
91.32 &
30.43 &
\multicolumn{1}{c|}{81.44} &
91.77 &
\textbf{1.02} &
\multicolumn{1}{c|}{\underline{96.37}} &
92.06 &
\underline{1.23} &
\multicolumn{1}{c|}{\textbf{96.41}} &
90.75 &
1.62 &
95.56 \\ \bottomrule
\end{tabular}%
\label{table1}
}
\end{table}

\begin{table}[h]
\caption{Comparison with the state-of-the-art defenses on Tiny ImageNet dataset with 5\% benign data and 10\% poison ratio on PreAct-ResNet18 (\%).}
\centering
\renewcommand\arraystretch{1.1}
\scalebox{0.65}{%
\begin{tabular}{cccccccccccccccc}
\toprule
\multicolumn{1}{c|}{\multirow{2}{*}{ATTACK}} &
\multicolumn{3}{c|}{Backdoored} &
\multicolumn{3}{c|}{FP \cite{liu2018fine}} &
\multicolumn{3}{c|}{NAD \cite{li2021neural}} &
\multicolumn{3}{c|}{NC \cite{wang2019neural}} &
\multicolumn{3}{c}{ANP \cite{wu2021adversarial}} \\
\multicolumn{1}{c|}{} &
ACC &
ASR &
\multicolumn{1}{c|}{DER} &
ACC &
ASR &
\multicolumn{1}{c|}{DER} &
ACC &
ASR &
\multicolumn{1}{c|}{DER} &
ACC &
ASR &
\multicolumn{1}{c|}{DER} &
ACC &
ASR &
DER \\ \midrule
\multicolumn{1}{c|}{BadNets-A2O \cite{gu2019badnets}} &
56.12 &
99.90 &
\multicolumn{1}{c|}{N/A} &
48.81 &
0.66 &
\multicolumn{1}{c|}{95.96} &
48.35 &
0.27 &
\multicolumn{1}{c|}{95.93} &
\textbf{56.12} &
99.90 &
\multicolumn{1}{c|}{50.00} &
47.34 &
\textbf{0.00} &
95.56 \\
\multicolumn{1}{c|}{BadNets-A2A \cite{gu2019badnets}} &
55.99 &
27.81 &
\multicolumn{1}{c|}{N/A} &
47.88 &
3.19 &
\multicolumn{1}{c|}{58.26} &
48.29 &
2.30 &
\multicolumn{1}{c|}{58.91} &
\underline{54.12} &
18.72 &
\multicolumn{1}{c|}{53.61} &
40.70 &
2.39 &
55.07 \\
\multicolumn{1}{c|}{Blended \cite{chen2017targeted}} &
55.53 &
97.57 &
\multicolumn{1}{c|}{N/A} &
50.58 &
57.89 &
\multicolumn{1}{c|}{67.37} &
\underline{55.22} &
98.88 &
\multicolumn{1}{c|}{49.84} &
54.50 &
96.07 &
\multicolumn{1}{c|}{50.24} &
43.21 &
43.80 &
70.73 \\
\multicolumn{1}{c|}{Input-Aware \cite{nguyen2020input}} &
57.67 &
99.19 &
\multicolumn{1}{c|}{N/A} &
52.38 &
0.13 &
\multicolumn{1}{c|}{96.88} &
\textbf{57.42} &
0.07 &
\multicolumn{1}{c|}{\textbf{99.44}} &
53.46 &
2.48 &
\multicolumn{1}{c|}{96.25} &
50.56 &
\textbf{0.00} &
96.04 \\
\multicolumn{1}{c|}{LF \cite{zeng2021rethinking}} &
55.21 &
98.51 &
\multicolumn{1}{c|}{N/A} &
48.18 &
63.83 &
\multicolumn{1}{c|}{63.82} &
49.61 &
58.01 &
\multicolumn{1}{c|}{67.45} &
53.08 &
90.48 &
\multicolumn{1}{c|}{52.95} &
41.75 &
65.98 &
59.54 \\
\multicolumn{1}{c|}{SSBA \cite{li2021invisible}} &
55.97 &
97.69 &
\multicolumn{1}{c|}{N/A} &
48.06 &
52.25 &
\multicolumn{1}{c|}{68.77} &
47.67 &
69.47 &
\multicolumn{1}{c|}{59.96} &
\underline{53.30} &
0.26 &
\multicolumn{1}{c|}{\textbf{97.38}} &
41.83 &
14.24 &
84.66 \\
\multicolumn{1}{c|}{Trojan \cite{Trojannn}} &
56.48 &
99.97 &
\multicolumn{1}{c|}{N/A} &
45.96 &
8.88 &
\multicolumn{1}{c|}{90.29} &
48.83 &
1.01 &
\multicolumn{1}{c|}{95.66} &
\underline{54.43} &
1.54 &
\multicolumn{1}{c|}{\underline{98.19}} &
45.36 &
0.53 &
94.16 \\
\multicolumn{1}{c|}{WaNet \cite{nguyen2021wanet}} &
57.81 &
96.50 &
\multicolumn{1}{c|}{N/A} &
50.35 &
1.37 &
\multicolumn{1}{c|}{93.84} &
50.02 &
0.87 &
\multicolumn{1}{c|}{93.92} &
\textbf{57.81} &
96.50 &
\multicolumn{1}{c|}{50.00} &
30.34 &
\textbf{0.00} &
84.52 \\ \midrule
\multicolumn{1}{c|}{Avg} &
56.35 &
89.64 &
\multicolumn{1}{c|}{N/A} &
49.03 &
23.53 &
\multicolumn{1}{c|}{79.40} &
50.68 &
28.86 &
\multicolumn{1}{c|}{77.55} &
\underline{54.60} &
50.74 &
\multicolumn{1}{c|}{68.58} &
42.64 &
15.87 &
80.03 \\ \bottomrule
\toprule
\multicolumn{1}{c|}{\multirow{2}{*}{ATTACK}} &
\multicolumn{3}{c|}{i-BAU \cite{zeng2022adversarial}} &
\multicolumn{3}{c|}{EP \cite{zheng2022pre}} &
\multicolumn{3}{c|}{NPD-TU (\textbf{Ours})} &
\multicolumn{3}{c|}{NPD-TP (\textbf{Ours})} &
\multicolumn{3}{c}{NPD (\textbf{Ours})} \\
\multicolumn{1}{c|}{} &
ACC &
ASR &
\multicolumn{1}{c|}{DER} &
ACC &
ASR &
\multicolumn{1}{c|}{DER} &
ACC &
ASR &
\multicolumn{1}{c|}{DER} &
ACC &
ASR &
\multicolumn{1}{c|}{DER} &
ACC &
ASR &
DER \\ \midrule
\multicolumn{1}{c|}{BadNets-A2O \cite{gu2019badnets}} &
51.63 &
95.92 &
\multicolumn{1}{c|}{49.74} &
\underline{54.00} &
0.02 &
\multicolumn{1}{c|}{\textbf{98.88}} &
47.23 &
\underline{0.01} &
\multicolumn{1}{c|}{95.50} &
49.89 &
1.28 &
\multicolumn{1}{c|}{\underline{96.20}} &
49.79 &
2.51 &
95.53 \\
\multicolumn{1}{c|}{BadNets-A2A \cite{gu2019badnets}} &
53.52 &
12.89 &
\multicolumn{1}{c|}{56.22} &
\textbf{54.79} &
\textbf{1.28} &
\multicolumn{1}{c|}{\textbf{62.67}} &
46.81 &
\underline{1.96} &
\multicolumn{1}{c|}{58.34} &
49.79 &
3.31 &
\multicolumn{1}{c|}{\underline{59.15}} &
49.94 &
5.57 &
58.10 \\
\multicolumn{1}{c|}{Blended \cite{chen2017targeted}} &
50.76 &
95.58 &
\multicolumn{1}{c|}{48.61} &
\textbf{56.32} &
88.88 &
\multicolumn{1}{c|}{54.34} &
46.24 &
\textbf{0.00} &
\multicolumn{1}{c|}{94.14} &
49.72 &
0.18 &
\multicolumn{1}{c|}{\textbf{95.79}} &
49.62 &
\underline{0.12} &
\underline{95.77} \\
\multicolumn{1}{c|}{Input-Aware \cite{nguyen2020input}} &
55.49 &
0.46 &
\multicolumn{1}{c|}{98.28} &
\underline{57.33} &
\underline{0.03} &
\multicolumn{1}{c|}{\underline{99.41}} &
49.54 &
0.27 &
\multicolumn{1}{c|}{95.40} &
53.88 &
0.04 &
\multicolumn{1}{c|}{97.68} &
53.75 &
5.93 &
94.67 \\
\multicolumn{1}{c|}{LF \cite{zeng2021rethinking}} &
\underline{53.65} &
94.27 &
\multicolumn{1}{c|}{51.34} &
\textbf{54.86} &
93.20 &
\multicolumn{1}{c|}{52.48} &
46.04 &
\textbf{0.00} &
\multicolumn{1}{c|}{94.67} &
49.20 &
\underline{0.30} &
\multicolumn{1}{c|}{\textbf{96.10}} &
49.94 &
2.48 &
\underline{95.38} \\
\multicolumn{1}{c|}{SSBA \cite{li2021invisible}} &
52.39 &
84.64 &
\multicolumn{1}{c|}{54.74} &
\textbf{55.56} &
66.67 &
\multicolumn{1}{c|}{65.31} &
46.56 &
\textbf{0.00} &
\multicolumn{1}{c|}{94.14} &
49.04 &
\underline{0.00} &
\multicolumn{1}{c|}{95.38} &
49.25 &
0.01 &
\underline{95.48} \\
\multicolumn{1}{c|}{Trojan \cite{Trojannn}} &
51.85 &
99.15 &
\multicolumn{1}{c|}{48.10} &
\textbf{54.47} &
0.12 &
\multicolumn{1}{c|}{\textbf{98.92}} &
48.56 &
\textbf{0.00} &
\multicolumn{1}{c|}{96.02} &
49.61 &
\underline{0.05} &
\multicolumn{1}{c|}{96.52} &
49.43 &
0.51 &
96.21 \\
\multicolumn{1}{c|}{WaNet \cite{nguyen2021wanet}} &
53.04 &
69.82 &
\multicolumn{1}{c|}{60.96} &
\underline{57.06} &
0.20 &
\multicolumn{1}{c|}{\textbf{97.78}} &
48.52 &
\underline{0.01} &
\multicolumn{1}{c|}{93.60} &
51.88 &
0.82 &
\multicolumn{1}{c|}{94.88} &
52.64 &
0.24 &
\underline{95.54} \\ \midrule
\multicolumn{1}{c|}{Avg} &
52.79 &
69.09 &
\multicolumn{1}{c|}{58.50} &
\textbf{55.55} &
31.30 &
\multicolumn{1}{c|}{78.77} &
47.44 &
\textbf{0.28} &
\multicolumn{1}{c|}{90.22} &
50.38 &
\underline{0.75} &
\multicolumn{1}{c|}{\textbf{91.46}} &
50.55 &
2.17 &
\underline{90.84} \\ \bottomrule
\end{tabular}%
\label{table2}
}
\end{table}

\subsection{Main results}
Table \ref{table1} and Table \ref{table2} showcase the defense performance of the proposed method in comparison to six SOTA defense methods on CIFAR-10 and Tiny ImageNet. The following observations can be made:

\begin{itemize}[leftmargin=*]
\item \textbf{Our methods show superior performance in terms of DER for almost all attacks compared to SOTA defenses.} Conversely, FP and ANP excel in maintaining high ACC, but they struggle to eliminate backdoors in strong attacks like Blended and LF. NC's emphasis on minimal universal adversarial perturbation renders it ineffective against sample-specific attacks and those utilizing large norm triggers. I-BAU shows similar performance in removing backdoors with an average DER of 92.42\%, but it leads to a significant decrease in ACC, likely due to training the entire network by adversarial training.
\item \textbf{Defense performance of NPD-TU and NPD-TP are better than NPD.} When the target label is known, the model only needs to find perturbations for that specific label, simplifying trigger identification and unlearning. These two methods outperform NPD, except for WaNet, which is a transformation-based attack without visible triggers. Fully perturbing the network proves more effective than solely unlearning targeted triggers in WaNet. 
\item  \textbf{NPD-TU is effective for trigger-additive attacks while NPD-TP is expert in defending against sample-specific attacks.} It can be observed by comparing defense results on different attacks like Blended and SSBA on CIFAR-10. This demonstrates that the applicability of different strategies varies across different attack scenarios. 
\item  \textbf{Defense performance is robust across all attacks on Tiny ImageNet.} Similar to the results on CIFAR-10, our method outperforms other methods in terms of ASR and DER for all backdoor attacks. Despite a slight decrease in ACC, NPD-TU achieving a remarkably good performance with ASR $<1.5\%$ on average. NPD-TP and NPD perform best in removing backdoors while maintaining model utility.
\end{itemize}

In summary, our method outperforms other state-of-the-art approaches, showcasing the broad applicability of our proposed method across diverse datasets. Due to space limits, defending results on GTSRB dataset and VGG19-BN network can be found in Section C of \textbf{supplementary materials}.

\begin{table}
\parbox{.49\linewidth}{

\caption{Defense performance under different components of losses.} \label{table4}
\renewcommand\arraystretch{1.1}
\scalebox{0.56}{%
\begin{tabular}{ccc|cc|cc|cc}
\toprule
\multicolumn{3}{c|}{ATTACK $\rightarrow$} & \multicolumn{2}{c|}{BadNets-A2O \cite{gu2019badnets}} & \multicolumn{2}{c|}{Blended \cite{chen2017targeted}} & \multicolumn{2}{c}{LF \cite{zeng2021rethinking}} \\
$\gL_{bce1}$ & $\gL_{bce2}$ & $\gL_{asr}$ & ACC   & ASR  & ACC   & ASR   & ACC   & ASR   \\ \midrule
&  &  & 91.45 & 1.18 & 92.47 & 99.63 & 92.00 & 95.90 \\
\checkmark &  &  & 90.17 & 1.19 & 91.51 & 2.01  & 90.91 & 9.60  \\
& \checkmark &  & 90.46 & 0.38 & 91.68 & 18.28 & 91.19 & 1.06  \\
&  & \checkmark & 90.02 & 0.27 & 91.31 & 98.32 & 91.07 & 0.80  \\
\checkmark & \checkmark &  & 89.56 & 0.21 & 91.09 & 1.73  & 90.47 & 7.63  \\
\checkmark & \checkmark & \checkmark & 88.93 & 1.26 & 91.18 & 0.41  & 90.06 & 0.21  \\ \bottomrule
\end{tabular}%

}}
\hspace{0.02em}
\parbox{.49\linewidth}{
\centering
\caption{Defense results in comparison with NPD-UU and NPD-UP.} \label{table3}
\renewcommand\arraystretch{1.1}
\vspace{0.2em}
\scalebox{0.56}{%
\begin{tabular}{c|cc|cc|cc|cc}
\toprule
\multirow{2}{*}{ATTACK $\downarrow$} & \multicolumn{2}{c|}{No defense} & \multicolumn{2}{c|}{NPD-UU} & \multicolumn{2}{c|}{NPD-UP} & \multicolumn{2}{c}{NPD (\textbf{Ours})} \\
& ACC   & ASR   & ACC   & ASR   & ACC   & ASR   & ACC   & ASR  \\ \midrule
BadNets-A2O \cite{gu2019badnets} & 91.82 & 93.79 & 79.35 & \bf{0.10}  & \bf{90.61} & 1.74  & 88.93 & 1.26 \\
Blended \cite{chen2017targeted}     & 93.44 & 97.71 & 86.35 & 10.77 & \bf{92.35} & 3.86  & 91.18 & \bf{0.41} \\
LF \cite{zeng2021rethinking}          & 93.01 & 99.06 & 82.95 & 75.42 & \bf{91.53} & 17.24 & 90.06 & \bf{0.21} \\
SSBA \cite{li2021invisible}        & 92.88 & 97.07 & 84.31 & 52.36 & \bf{91.49} & 14.22 & 90.88 & \bf{2.77} \\
Trojan \cite{Trojannn}      & 93.47 & 99.99 & 89.81 & 38.11 & \bf{92.61} & 11.43 & 92.37 & \bf{6.51} \\
WaNet \cite{nguyen2021wanet}       & 92.80 & 98.90 & 84.70 & 5.98  & \bf{92.11} & 1.41  & 91.57 & \bf{0.80} \\ \bottomrule
\end{tabular}%

}}

\end{table}

\subsection{Analysis}
\paragraph{Effectiveness of each loss term.}
We conduct an ablation study to evaluate the contribution of each component of the loss function towards the overall performance on CIFAR-10 dataset. We separately investigate the first and second terms of loss $\gL_{bce}$ (see Eq. (\ref{eq: bce loss})), denoting them as $\gL_{bce1}$ and $\gL_{bce2}$, respectively. Throughout the study, we keep the loss $\gL_{bn}$ consistent across all experiments and the result is shown in Table \ref{table4}. Notably, the loss $l_{bce1}$ plays a significant role in improving the overall performance, while removing each component leads to a significant drop in defense in certain cases. This ablation study underscores the importance of each loss component in effectively mitigating different types of attacks.

\paragraph{Effectiveness of the targeted adversarial perturbations in NPD.} 
To show the efficacy of the targeted adversarial perturbations in NPD (see Eq. (\ref{loss})), we compare NPD with its two variants using two types of untargeted perturbation. 
We refer to adversarial perturbations generated by UAP and standard PGD without a targeted label as NPD-UU (untargeted  universal adversarial perturbation) and NPD-UP (untargeted PGD), respectively. As shown in Table \ref{table3}, NPD-UU and NPD-UP fail to remove backdoors in certain cases although NPD-UP obtains a higher ACC. This result shows the superiority of NPD in removing backdoors.

\begin{wrapfigure}{r}{6.5cm}
\centering
\includegraphics[width=\linewidth]{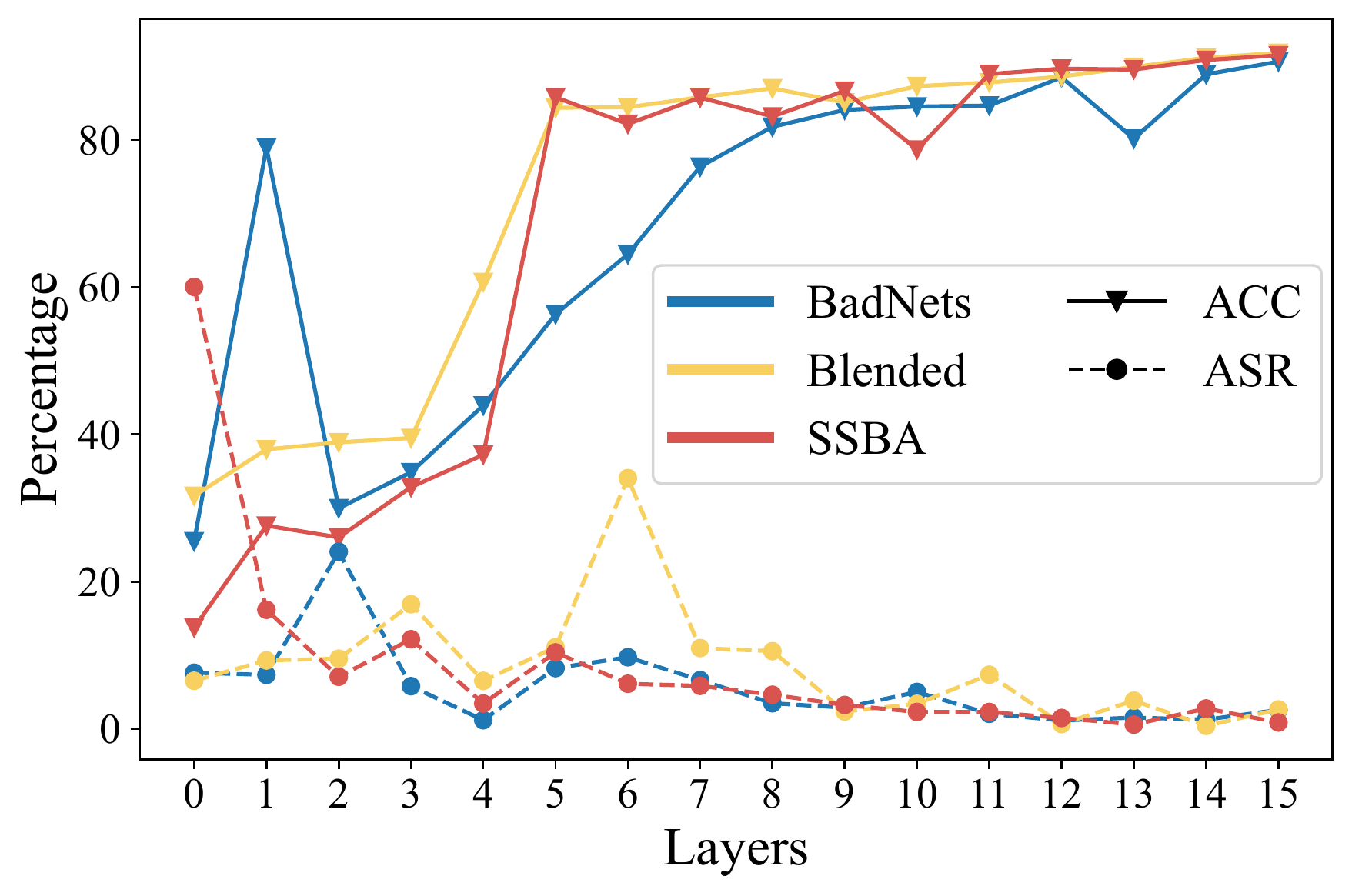}
\caption{Defense performance of inserting linear transformation into different layers.}
\vspace{-2em}
\label{fig:layers}
\end{wrapfigure}

\paragraph{Performance of choosing different layers to insert the transformation layer.}
We evaluate the influence of choosing different layers to insert the transformation layer by inserting it after each convolution layer of PreAct-ResNet18 network on CIFAR-10 dataset. Figure \ref{fig:layers} shows the defense performance under three attacks. The result shows that inserting the transformation layer into the shallower layers results in a decrease in accuracy. This is because even a slight perturbation in the shallow layers can cause significant instability in the final output. However, as the layer goes deeper, the features become more separable, resulting in better defense performance.

\paragraph{Defense effectiveness under different poisoning ratios.}
\begin{table}[!t]
\caption{Defense results under different poisoning ratio on CIFAR-10 and PreAct-ResNet18(\%).}
\centering
\renewcommand\arraystretch{1.1}
\setlength{\tabcolsep}{4pt}
\scalebox{0.674}{%
\begin{tabular}{cc|cc|cc|cc|cc|cc}
\toprule
\multicolumn{2}{c|}{Poisoning Ratio $\rightarrow$} &
\multicolumn{2}{c|}{5\%} &
\multicolumn{2}{c|}{10\%} &
\multicolumn{2}{c|}{20\%} &
\multicolumn{2}{c|}{30\%} &
\multicolumn{2}{c}{40\%} \\
\multicolumn{2}{c|}{ATTACK $\downarrow$} &
No Defense &
Ours &
No Defense &
Ours &
No Defense &
Ours &
No Defense &
Ours &
No Defense &
Ours \\ \midrule
\multirow{2}{*}{BadNets-A2O \cite{gu2019badnets}} & ACC & 92.35 & 87.99 & 91.82 & 88.93 & 90.17 & 85.77 & 88.32 & 86.76 & 86.16 & 82.31 \\
& ASR & 89.52 & 0.89  & 93.79 & 1.26  & 96.12 & 0.39  & 97.33 & 0.69  & 97.78 & 3.88  \\ \midrule
\multirow{2}{*}{Blended \cite{chen2017targeted}}     & ACC & 93.76 & 91.48 & 93.44 & 91.18 & 93.00 & 91.27 & 92.78 & 90.10 & 91.64 & 88.85 \\
& ASR & 99.31 & 11.06 & 97.71 & 0.41  & 99.92 & 4.41  & 99.98 & 3.78  & 99.96 & 24.92 \\ \midrule
\multirow{2}{*}{Input-Aware \cite{nguyen2020input}}  & ACC & 90.92 & 90.31 & 94.03 & 89.57 & 89.18 & 88.60 & 89.63 & 89.41 & 90.12 & 88.57 \\
& ASR & 94.19 & 0.30  & 98.35 & 0.11  & 97.66 & 3.51  & 97.62 & 3.43  & 98.57 & 0.98  \\ \midrule
\multirow{2}{*}{WaNet \cite{nguyen2021wanet}}       & ACC & 93.38 & 91.38 & 92.80 & 91.57 & 91.02 & 89.77 & 92.35 & 89.53 & 92.21 & 89.53 \\
& ASR & 97.27 & 0.14  & 98.90 & 0.80  & 94.93 & 0.22  & 99.06 & 1.71  & 99.49 & 2.19  \\ \bottomrule
\end{tabular}%
\label{table5}
}
\vspace{-1em}
\end{table}

To investigate the impact of poisoning ratios on defense performance, we conducted experiments on NPD with different poisoning ratios on the CIFAR-10 dataset. As presented in Table \ref{table5}, there is a slight decrease in ACC as the poisoning ratio increases. Moreover, our approach exhibits a notably stable defense performance across a range of poisoning ratios.

\paragraph{Defense effectiveness under different clean ratios.}


\begin{table}[h]
\caption{Results with different number of clean data on CIFAR-10 (\%). The best DERs are highlighted in \textbf{Boldface}.}
\renewcommand\arraystretch{1.1}
\centering
\scalebox{0.674}{%
\begin{tabular}{c|c|ccc|ccc|ccc}
\toprule
\multirow{2}{*}{ATTACK}      & $\mathbf{\#}$ Clean data $\rightarrow$ & \multicolumn{3}{c|}{500} & \multicolumn{3}{c|}{250} & \multicolumn{3}{c}{50} \\
& Defense $\downarrow$ & ACC   & ASR   & DER   & ACC   & ASR   & DER   & ACC   & ASR   & DER   \\ \midrule
\multirow{3}{*}{BadNets-A2O \cite{gu2019badnets}} & i-BAU \cite{zeng2022adversarial}        & 65.41   & 7.74  & 79.82  & 66.05   & 2.02  & 81.03  & 64.60  & 22.10 & 70.27 \\
& ANP \cite{wu2021adversarial}     & 91.53 & 5.50  & 94.00 & 90.81 & 2.03  & \textbf{93.40} & 85.57 & 1.52  & 91.04 \\
& Ours    & 87.58 & 1.38  & \textbf{94.09} & 87.80 & 0.27  & 92.78 & 86.91 & 0.20  & \textbf{92.37} \\ \midrule
\multirow{3}{*}{SSBA \cite{li2021invisible}} & i-BAU \cite{zeng2022adversarial}   & 84.21 & 17.73 & 85.33 & 76.84 & 51.90 & 60.38 & 67.48 & 98.37 & 35.21 \\
& ANP \cite{wu2021adversarial}     & 92.06 & 28.67 & 83.79 & 92.13 & 27.61 & 80.17 & 88.31 & 22.28 & 80.92 \\
& Ours    & 90.48 & 0.43  & \textbf{97.12} & 90.80 & 10.03 & \textbf{88.29} & 89.48 & 8.97  & \textbf{88.16} \\ \midrule
\multirow{3}{*}{LF \cite{zeng2021rethinking}}   & i-BAU \cite{zeng2022adversarial}   & 83.91 & 19.34 & 85.31 & 70.12 & 99.42 & 35.53 & 70.26 & 99.31 & 35.60 \\
& ANP \cite{wu2021adversarial}     & 92.74 & 46.70 & 76.04 & 92.28 & 18.02 & 84.11 & 89.99 & 18.77 & 82.59 \\
& Ours    & 90.21 & 0.99  & \textbf{97.63} & 89.28 & 0.50  & \textbf{91.37} & 88.18 & 10.83 & \textbf{85.65} \\ \bottomrule
\end{tabular}%
\label{table7}
}
\end{table}

We investigate the sensitivity of clean data on defense performance and compare our NPD with SOTA defenses. As shown in Table \ref{table7}, NPD is less sensitive to the size of clean data among all the attacks and defenses. Even with only 50 samples, it still maintains acceptable performance. This result shows that our method exhibits minimal reliance on the number of training samples.

\paragraph{Running time comparison.}

\begin{table}[h]
\caption{Running time of different defense methods with 2500 CIFAR-10 images on PreActResNet18.}
\renewcommand\arraystretch{1.1}
\centering
\setlength{\tabcolsep}{4pt}
\scalebox{0.8}{%
\begin{tabular}{c|ccccccc}
\toprule
Defense            & FP \cite{liu2018fine}      & NAD \cite{li2021neural}   & NC \cite{wang2019neural}     & ANP \cite{wu2021adversarial}   & i-BAU \cite{zeng2022adversarial} & EP \cite{zheng2022pre}    & NPD (\textbf{Ours}) \\ \midrule
Runnign Time (sec.) & 1169.01 & 74.39 & 896.45 & 58.75 & 57.23 & 131.84 & 55.16  \\ \bottomrule
\end{tabular}%
\label{table6}
}
\end{table}

We measure the runtime of the defense methods on 2500 CIFAR-10 images with batch size 256 and PreAct-ResNet18. The experiments were conducted on a RTX 4090Ti GPU and the results are presented in Table \ref{table6}. Among these methods, our proposed PND-UN achieves the fastest performance, requiring only 56 seconds. It should be noted that our method was trained for 50 epochs, while i-BAU was only trained for 5 epochs.

\section{Conclusion}
Inspired by the mechanism of optical polarizer, this work proposed a novel backdoor defense method by inserting a learnable neural polarizer as an intermediate layer of the backdoored model. We instantiated the neural polarizer as a lightweight linear transformation and it could be efficiently and effectively learned with limited clean samples to mitigate backdoor effect. 
To learn a desired neural polarizer, a bi-level optimization problem is proposed by filtering trigger features of poisoned samples while maintaining benign features of both poisoned and benign samples. Extensive experiments demonstrate the effectiveness of our method across all evaluated backdoor attacks and all other defense methods under various datasets and network architectures. 

\paragraph{Limitations and future work.} Although only limited clean data is needed for our method to achieve a remarkable defense performance, the accessibility of clean data is still an important limitation of the proposed method, which may restrict the application of our method. Therefore, a promising direction for future work is to further reduce the requirement of clean data by exploring data-free neural polarizer or learning neural polarizer based on poisoned training data.

\paragraph{Broader impacts.} Backdoor attacks pose significant threats to the deployment of deep neural networks obtained from untrustworthy sources. 
This work has made a valuable contribution to the community with an efficient and effective backdoor defense strategy to ease the threat of existing backdoor attacks, even with a very limited set of clean samples, which ensures its practicality.  
Besides, the innovative defense strategy of learning additional lightweight layers, rather than adjusting the whole backdoored model, may inspire more researchers to develop more efficient and practical defense methods.

\clearpage
{\small
\bibliography{main}
}

\end{document}